\documentclass[10pt,twocolumn]{article} 
\usepackage{simpleConference}
\usepackage{times}
\usepackage{graphicx}
\usepackage{amssymb}
\usepackage{url,hyperref}
\usepackage[utf8]{inputenc} % allow utf-8 input
\usepackage[T1]{fontenc}    % use 8-bit T1 fonts
\usepackage{booktabs}       % professional-quality tables
\usepackage{amsfonts}       % blackboard math symbols
\usepackage{nicefrac}       % compact symbols for 1/2, etc.
\usepackage{graphicx}
\usepackage{multirow}
\usepackage{tikz}
\usetikzlibrary{shapes.geometric, arrows}
\usetikzlibrary{fit,backgrounds} 
\usepackage{multicol}
\usepackage{pifont}
\usepackage{makecell}
\usepackage{svg}
\usepackage{amsmath}
\usepackage{amssymb}
\usepackage{float}
\usepackage{url}
\usepackage{xcolor}
\usepackage{soul}
\usepackage{booktabs}

\newcommand{\eg}{\textit{e.g.}}

\begin{document}
\title{\LARGE \bf A Generative Approach Towards Improved\\ Robotic Detection of Marine Litter}

\author{Jungseok Hong$^{1}$, Michael Fulton$^{2}$, and Junaed Sattar$^{3}$
\thanks{The authors are with the Department of Computer Science and Engineering, Minnesota Robotics Institute, University of Minnesota--Twin Cities, 100 Union St SE, Minneapolis, MN, 55455, USA
{\tt\small \{$^{1}$jungseok, $^{2}$fulto081, $^{3}$junaed\} at umn.edu.}}
}
% \author{Iroro Orife \\
% \\
% Technical Report \\
% Seattle, Washington, USA \\
% \today
% \\
% \\
% iroro@alumni.cmu.edu  \\
% }

\maketitle
\thispagestyle{empty}

\begin{abstract}
This paper presents an approach to address data scarcity problems in underwater image datasets for visual detection of marine debris. The proposed approach relies on a two-stage variational autoencoder (VAE) and a binary classifier to evaluate the generated imagery for quality and realism. From the images generated by the two-stage VAE, the binary classifier selects ``good quality'' images and augments the given dataset with them. Lastly, a multi-class classifier is used to evaluate the impact of the augmentation process by measuring the accuracy of an object detector trained on combinations of real and generated trash images. Our results show that the classifier trained with the augmented data outperforms the one trained only with the real data. This approach will not only be valid for the underwater trash classification problem presented in this paper, but it will also be useful for any data-dependent task for which collecting more images is challenging or infeasible.
% Latest advances in neural networks have introduced breakthrough innovations in computer vision and robotics. However, advances in underwater perception, particularly for autonomous robotic vehicles, have been relatively scarce because of the vagaries of the domain itself and the lack of available data. 
% %Augmenting datasets simply with flipped and rotated images is not an sufficient approach to add diverse variability. 
% In this paper, we propose an approach to expand our existing dataset for underwater trash detection and classification by using generative models. We first adopt a two-stage VAE model to generate realistic underwater images. Then, the ``good'' generated images are selected by a binary classifier and added to the datasets. Lastly, the impact of the expanded dataset is evaluated with a multi-class classifier for two classes of trash: plastic bags and bottles. The results demonstrate that the proposed approach is effective to handle data scarcity problems when it is infeasible to collect any more data, as in our application of underwater trash classification.
\end{abstract}

\section{Introduction}
\label{sec:introduction}
% Underwater robotics has experienced significant development in recent years. This has led some development in research. Due to the progress in computer vision domain, it helped to improve underwater object detection and classification. However, deep learning based image classification requires a lot of data b
% Generative model is good since underwater images are harder to collect than images from ground due to the restrictions. (specialized human is required). This is the perfect place that we can apply generative model so it can generate images of objects that can be found underwater. As it becomes more domain specific, it is even hard to collect data.
% Generative models have become popular due to the success of GAN. It includes, a lot of effort to tune hyperparameters.
% Over the past few years, many methods[r] have been explored in order to give an Autonomous Underwater vehicles (AUV) the ability to understand and infer its environment. Many of these methods rely on data from sensors such as Light Detection and Ranging (LIDAR), Laser rangefinder, Sound Navigation And Ranging (SONAR), and camera. Among the sensors, a camera is an attractive option due to its non-intrusive and passive nature, as well as being an intuitive medium for humans. 
Recent developments in computer vision, particularly spurred on by advances in deep convolutional neural networks (CNNs), have greatly benefited mobile robotics in applications such as autonomous driving \cite{chen2017multi}, drone surveillance \cite{guvenc2018detection}, harvesting machines \cite{zhang2018deep}, and milking robots \cite{wisse2018milking}. However, machine vision in the underwater domain has not received similar attention compared to its terrestrial counterpart, limiting use of vision-guided methods for real-world problems. As an example, visual detection of underwater trash~\cite{fulton2018robotic} by autonomous underwater vehicles (AUVs) suffers from lower accuracy compared to the performance of the same deep-learned detectors in non-underwater object detection tasks. The training dataset, which is significantly smaller than the size of other non-underwater training datasets, is likely a contributing factor for this. Significantly expanding the dataset would be a preferable approach to improve trash detection performance; however, collecting imagery of marine debris is not just time consuming, costly, and physically demanding, in many cases it is infeasible due to the sheer logistics involved in underwater operations. Although methods such as few-shot learning \cite{garcia2017few}, zero-shot learning \cite{palatucci2009zero}, and transfer learning \cite{torrey2010transfer} have been used to address data scarcity issues, initial experiments with transfer learning in the author's previous work~\cite{fulton2018robotic} did not show notable improvements. 
\begin{figure}[t!]
\setlength{\lineskip}{0pt}
\centering
\setlength\tabcolsep{1.pt}
\renewcommand{\arraystretch}{0.5}
  \begin{tabular}{cc}
    \includegraphics[trim=0 0 0 0, width=.245\textwidth]{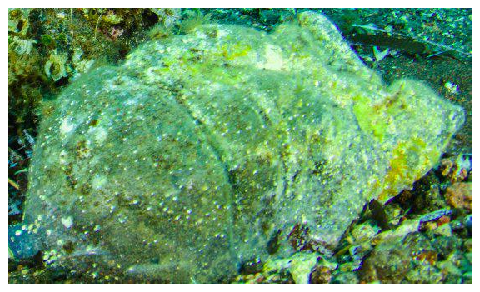} &
    \includegraphics[trim=0 0 0 5.5,clip,width=.23\textwidth]{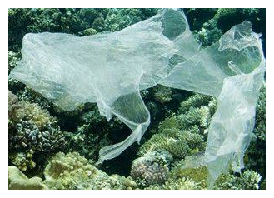} \\
    \includegraphics[trim=0 0.1 0 0,clip,width=.249\textwidth]{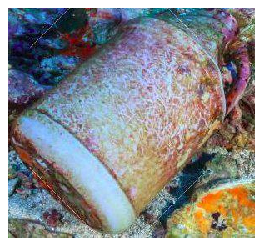} &
    \includegraphics[trim=0 0.1 0 0,clip,width=.23\textwidth]{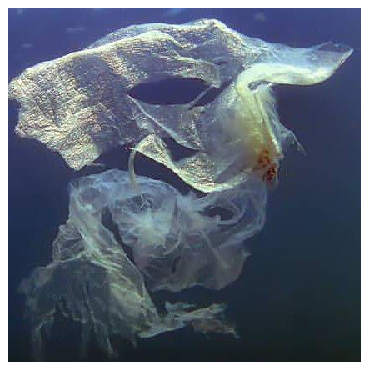} \\
  \end{tabular}
  \caption{Sample images showing real underwater trash. Objects, made of plastic and metal, are at different stages of shape and color deformation.}
  \label{fig:introim}
\end{figure}

\begin{figure*}[t!]
\setlength{\lineskip}{0pt}
\centering
\setlength\tabcolsep{1.pt}
\renewcommand{\arraystretch}{0.5}
  \begin{tabular}{cccc}
    \includegraphics[width=.278\textwidth]{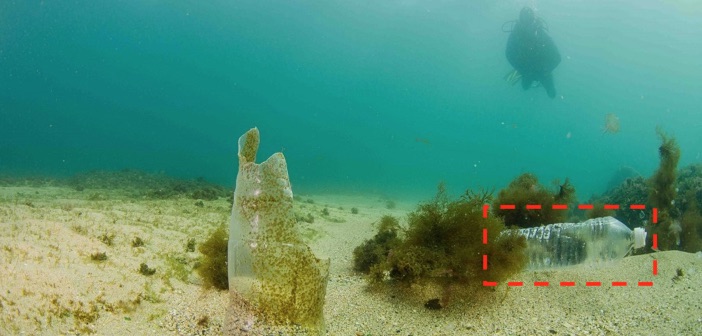} &
    \includegraphics[width=.235\textwidth]{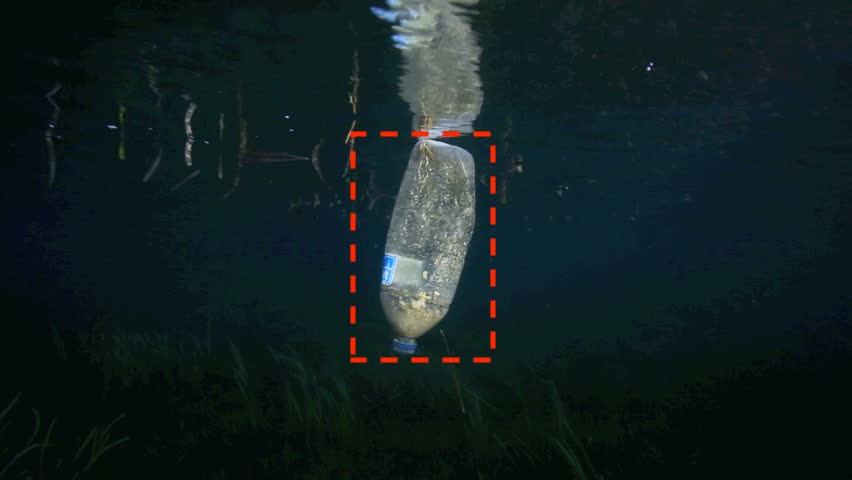} &
    \includegraphics[width=.235\textwidth]{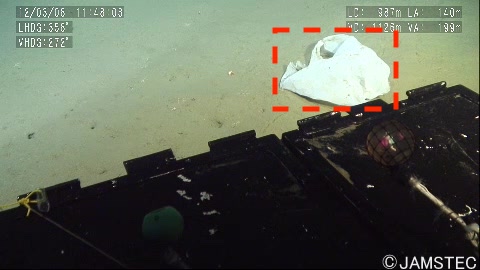} &
    \includegraphics[width=.235\textwidth]{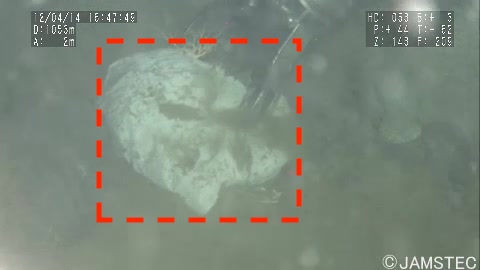}    \\
  \end{tabular}
  \caption{Examples of underwater trash images collected and annotated for training the proposed model. Images are collected from a variety of underwater conditions, and contain a variety of objects ranging from plastics to metals.}
  \label{fig:collec}
\end{figure*}

Numerous other underwater domain-specific issues also add to the challenge; \eg, (1) the shapes of objects degrades with long-term water exposure, (2) color absorption increases with depth, (3) light scatters and refracts underwater, causing images to become blurred and distorted. 
%After 20-meter depth, only green and blue color can be seen, and after 80-meter depth, no visible light exists 
Fig.~\ref{fig:introim} shows some real-world underwater trash images of plastic bags and plastic bottles. As can be seen, the shape of the objects in the images and also their colors are distorted and appears drastically different from when they would be seen over-water, not to mention the blurred overall appearance. These examples illustrate the issues mentioned above, which cannot simply be handled by commonly-used image augmentation techniques~\cite{perez2017effectiveness}. Because of variations in the shape of objects underwater (particularly predominant in aquatic trash), domain transfer~\cite{li2017watergan,  cao2018recent} on non-underwater images is unlikely to produce significantly realistic underwater imagery to expand a dataset for training. Also, domain transfer methods will not be able to capture the surrounding marine flora and fauna where debris is like to be found. While data augmentation techniques will produce a larger dataset, it may not be effective for training an object detector as it may fail to increase dataset variation.

In this paper, we focus on improving underwater object classification tasks to enable highly accurate trash detection by proposing a novel approach to realistically extend available datasets both in volume and variety. Specifically, we adopt deep generative models, in particular variational autoencoders, to synthesize realistic images of underwater trash. Generative methods estimate a probabilistic mapping from random noise sampled from a Gaussian distribution to input data (\eg, images). After training, they are used to generate realistic representations of the input data by feeding random noise to the mapping.
In this way, we can increase the number of images while adding variety by using a generative model to create synthetic images to expand the original dataset. 
Generative models have gained popularity since the appearance of Generative Adversarial Network (GAN) \cite{goodfellow2014generative} due to its relatively straightforward and effective approaches to tackle the generation problem. Other generative models such as Variational Autoencoder (VAE) \cite{kingma2013auto}, Bayesian networks \cite{friedman1997bayesian}, and Gaussian mixture models \cite{zivkovic2004improved} have also been shown to work well in image generation problems. We have decided to use VAE-based models to generate synthetic images, for reasons which are discussed in Section~\ref{sec:related_work}.

% Generative models estimate a probabilistic model which maps latent variables to synthetic images. 
% \missingfigure[figwidth=6cm]{Testing a long text string}
% \noindent\includegraphics[scale=0.75]{example-image-a} 

In this paper, we make the following contributions:
\begin{itemize}
    \item We propose an approach to expand an existing small dataset of underwater trash for training object detection and classification models.
    \item As an example of that approach, we develop a generative model using a two-stage VAE with a small-size training dataset to generate synthetic trash images.
    \item We evaluate the success of this approach on underwater object classification tasks by training detectors on purely real, purely generated, and mixed (real augmented with generated) data. 
\end{itemize}

% \hl{as you can see in the images, trash does not hold its original shape, it gets decayed due to chemical response. Therefore, it is necessary to do more than just chaning color and this is where generative model can play a key role.}

\section{Related Work}
\label{sec:related_work}

% Due the the significant research development in generative models in 

Various generative models \cite{wang2019generative} have been proposed to generate realistic imagery, and they are used for data augmentation in the non-underwater domain \cite{antoniou2017data} to improve object classification and detection tasks. Although there are generative model-based studies for the underwater domain, they mainly focus on image enhancements, such as WaterGAN \cite{li2017watergan}, enhancing underwater imagery using generative adversarial networks (UGAN) \cite{fabbri2018enhancing}, and the synthesis of unpaired Underwater images using a Multistyle Generative Adversarial Network (UMGAN) \cite{cao2018recent}. 

WaterGAN uses pairs of non-underwater images and depth maps to train its generative model. Then, it generates underwater images from above-water images. UGAN uses CycleGAN \cite{zhu2017unpaired} to generate distorted images from images with no distortion, with the Wasserstein GAN \cite{arjovsky2017wasserstein} used to prevent the adversarial training process from destabilizing. UMGAN combines CycleGAN and Conditional GAN \cite{mirza2014conditional} to generate multistyle underwater images. Although UMGAN generates images with various turbidity and colors, output images are equivalent to the input images except for their colors and turbidity. With all these methods, it is challenging to realistically simulate color and shape distortions of various kinds of materials underwater since they have no effect on the objects which are present in the original image; these methods simply perform a broad domain-transfer technique to simulate underwater imagery. 

 %are widely adapted and have gained popularity for generating images with sharper visual features than other generative models. 
 
 It can be challenging to tune hyperparameters for GAN models, which aim to find the Nash equilibrium to a two player min-max problem \cite{brock2018large, salimans2016improved}. While VAE-based approaches generate blurrier images than GAN-based ones, those based on the VAE are more stable during training and less sensitive to hyperparameter choices. Among recent VAE enhancements, only the two-stage VAE \cite{dai2019diagnosing} focuses on improving visual quality scores rather than improving log-likelihood scores. As a result, the two-stage VAE produces high-fidelity images which have visual quality scores close to the GAN enhancements while being much more stable during training.

As mentioned earlier in this section, the approaches adding various effects to input images are limited in that they can only create images they have seen. This is not effective to solve data scarcity problems in the underwater domain. Therefore, it is necessary to develop generative models which generate underwater images that are not tied to input images.

\section{Methodology}
\label{sec:methodology}

\begin{figure}[h]
\begin{center}
\begin{tikzpicture}[node distance=2cm]
\tikzstyle{startstop} = [rectangle, rounded corners, minimum width=3cm, minimum height=1cm,text centered, draw=black, fill=red!30]
\tikzstyle{io} = [trapezium, trapezium left angle=70, trapezium right angle=110, minimum width=3cm, minimum height=1cm, text centered, draw=black, fill=blue!30]
\tikzstyle{process} = [rectangle, minimum width=3cm, minimum height=1cm, text centered, draw=black, fill=orange!30]
\tikzstyle{process2} = [rectangle, minimum width=3cm, minimum height=1cm, text centered, draw=black, fill=green!30]
\tikzstyle{decision} = [diamond, minimum width=3cm, minimum height=1cm, text centered, draw=black, fill=green!30]
\tikzstyle{arrow} = [thick,->,>=stealth]
% \node (start) [startstop] {Start};
\node (pro1) [process2] {Underwater two-stage VAE};
\node (dec1) [decision, below of=pro1, yshift=-1cm] {Binary classifier};
\node (pro2a) [process, below of=dec1, yshift=-1cm] {Append images to dataset};
\node (pro2b) [process, right of=dec1, xshift=2.7cm] {Discard images};

\draw [arrow] (pro1) -- node[anchor=west] {Generate images} (dec1);
\draw [arrow] (dec1) -- node[anchor=west] {Good images} (pro2a);
\draw [arrow] (dec1) -- ++(3.2cm, 0) node[anchor=south, xshift=-0.9cm]{Bad images} (pro2b);

\end{tikzpicture}
\end{center}
\caption{Flowchart showing image generation, filtering and augmentation process for expanding an image dataset (color green represents trained models).} \label{fig:flo}
\end{figure}
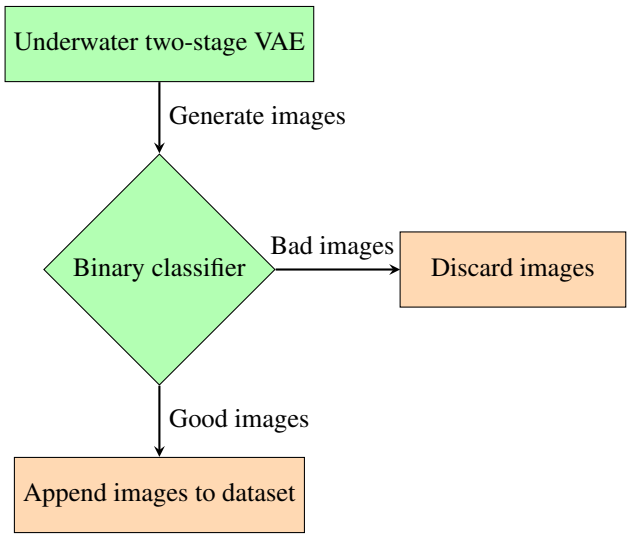

\begin{figure*}
    \centering
    \includegraphics[width=.99\textwidth]{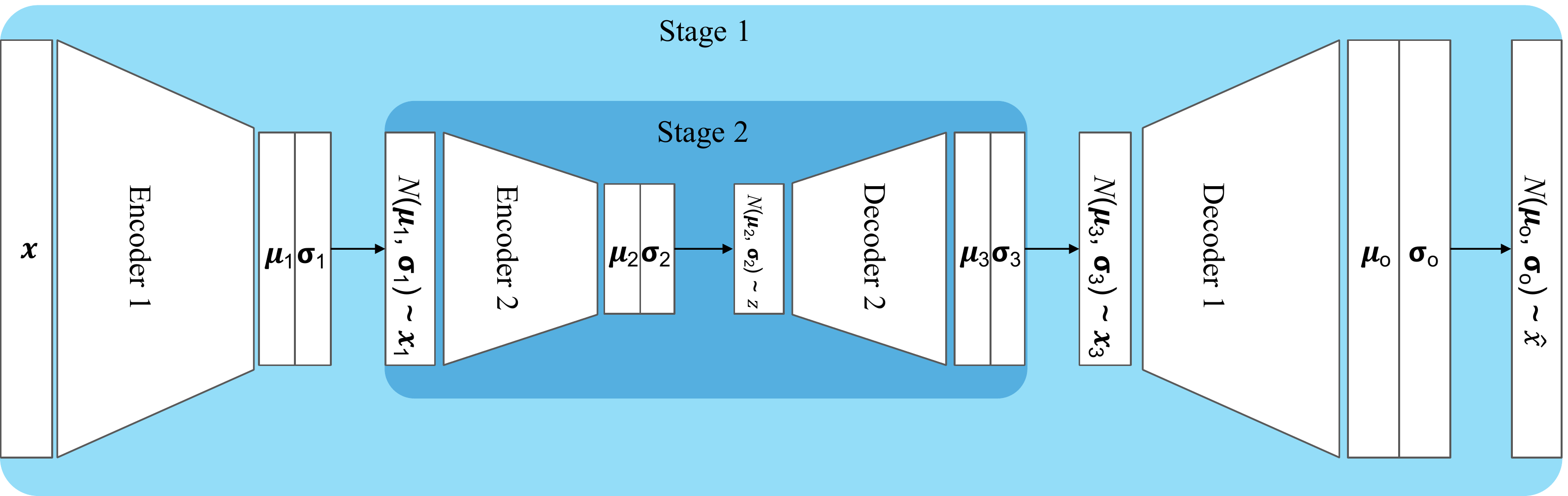}
    \caption{The architecture of the two-stage VAE.}
    \label{fig:twostage}
\end{figure*}
\begin{figure*}[h]
\centering
    \includegraphics[width=.9\textwidth]{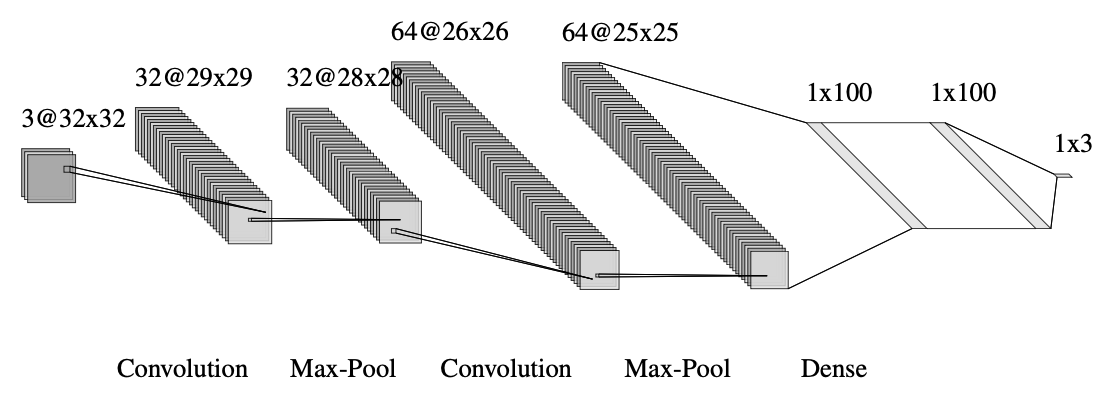}
    % \includesvg[width=.8\textwidth]{images/nn.svg}
    \caption{The architecture of the multi-class classifier.}
    \label{fig:nn}
\end{figure*}

The proposed two-stage VAE model is trained with an existing corpora of underwater trash images (see Section~\ref{sec:experiments} for details). After training, the model is used to generate synthetic underwater trash images, which are subsequently classified by hand (human observers) as ``good'' or ``bad'' based on their quality. This generated and classified data is then used to train an automated binary classifier. After training both the VAE and the binary classifier, we are able to generate and append good quality images to our dataset; the process is depicted in Fig. \ref{fig:flo}. Lastly, the effect of the dataset expansion is evaluated with a multi-class classifier trained to discriminate between trash and non-trash objects.

\subsection{Data Collection}
We collect and annotate images of plastic bags and plastic bottles since these are common objects which have a wide variety of shapes and frequently deform significantly over time underwater; also, plastic makes up the lion's share of marine debris and has the most detrimental effect on the ecosystem. This makes it challenging to build a truly representative dataset purely from observations. The images are obtained from the J-EDI (JAMSTEC E-Library of Deep-sea Images) dataset~\cite{JAMSTECDebri} and web scraping, which are then cropped to have tight bounds around the contained objects as shown in Fig~\ref{fig:collec}. The J-EDI dataset consists of images which have been collected since 1982 at various locations and depths. It is also sorted by types of trash. We collect plastic trash video clips from the J-EDI dataset, label objects in each clip, and build our own dataset, which will be released in the near future. Lastly, we add images from the web to bring more variety to our dataset.

\subsection{Variational Autoencoder}
The VAE aims to learn a generative model of input data $x \in \mathcal{X}$. The dataset $\mathcal{X}$ is $d$-dimensional and it has a $r$-dimensional manifold. The value of $r$ is unknown. We use the term `VAE' to denote a single-stage VAE in the rest of this paper.
\subsubsection{Single-stage VAE (original VAE)}
\label{vaeex}
The VAE consists of two neural networks which are an encoder $q_{\phi}(z|x)$ and a decoder $p_{\theta}(x|z)$. The encoder outputs the parameters of the normal distribution, $\mu$ and $\sigma$. From $N(\mu, \sigma)$, the $\kappa$-dimensional latent variable $z$ is sampled. Once the latent variable $z$ is sampled and provided as an input to the decoder, it generates the reconstructed original input $\hat{x}$.
The cost function for the VAE is shown in Eq. \ref{eq:loss} where $\mu_{gt}$ is a ground-truth probability measure and $\int_{\mathcal{X}}\mu_{gt}(dx)=1$. The function is minimized by stochastic gradient descent and the VAE jointly learns the latent variable and inference models during the training.
\begin{equation}\label{eq:loss}
\begin{split}
    \mathcal{L(\theta, \phi)} = \int_\mathcal{X}\Big\{-\mathbb{E}_{z\sim q_{\phi}(z|x)}[\log p_{\theta}(x|z)]\\
    +\mathbb{KL}[q_{\phi}(z|x)||p(z)]\Big\}\mu_{gt}(dx)
\end{split}
\end{equation}

\subsubsection{Two-stage VAE}
The VAE, compared to the GAN, (1) gives interpretable networks \cite{brock2016neural}, and (2) remains stable during training~\cite{tolstikhin2017wasserstein}. However, the Gaussian assumption for the encoder and decoder of the VAE reduces the quality of generated images~\cite{kingma2016improved}. 

In \cite{dai2019diagnosing}, the authors show that there exist the parameters $\phi$ and $\theta$ which optimize the VAE objective with the Gaussian assumptions. Based on this finding, the authors show that the two-stage VAE finds the parameters $\phi$ and $\theta$ closer to the optimal solution than the parameters found by the VAE. The structure of the two-stage VAE is shown in Fig. \ref{fig:twostage}. As explained in section \ref{vaeex}, each encoder yields parameters for the normal distribution and the values sampled from the distribution are fed to each decoder to generate the reconstructed input. 

The cost function in Eq. \ref{eq:loss} is used without any modification for the two-stage VAE. Each stage of the two-stage VAE is trained sequentially rather than jointly from the first to the second stage. We chose the ResNet structure \cite{he2016deep} for the stage 1 VAE, and selected four fully connected (FC) layers for the stage 2 VAE. Each FC layer is $1024$-dimensional in our model. The network for the stage 2 VAE is smaller than the stage 1 VAE since it is assumed that $d \gg \kappa \geq r$.

We use Fr\'{e}chet Inception Distance (FID) score \cite{heusel2017gans} to evaluate the quality of generated images. Lower FID scores represent higher image quality. It extracts features from generated images and real images using an intermediate layer of InceptionNet~\cite{szegedy2015going}. The data distribution of the features from generated images and real images are modeled separately using a multivariate Gaussian distribution. Then the FID score is calculated by Eq. \ref{eq:fid} where $\mu$ is mean, and $\Sigma$ is covariance. 
% Originally, the Inception Score (IS) was used but it is more sensitive to noise than the FID score. We use the FID score to evaluate the quality of generated images.
\begin{equation}
    \mathbf{FID}(x,g) = ||\mu_{x} - \mu_{g}||^{2}_{2} + Tr(\Sigma_{x} + \Sigma_{g} -2(\Sigma_{x}\Sigma_{g})^{\frac{1}{2}})
    \label{eq:fid}
\end{equation}

\begin{figure*}[t!]
\setlength{\lineskip}{0pt}
\centering
\setlength\tabcolsep{1.pt}
\renewcommand{\arraystretch}{0.5}
  \begin{tabular}{c@{\extracolsep{0.1cm}}c@{\extracolsep{0.1cm}}c@{\extracolsep{0.1cm}}c@{\extracolsep{0.1cm}}c@{\extracolsep{0.5cm}} c @{\extracolsep{0.1cm}}c@{\extracolsep{0.1cm}}c@{\extracolsep{0.1cm}}c}
  & \multicolumn{4}{c}{Plastic Bag} & \multicolumn{4}{c}{Plastic Bottle} \\
  \raisebox{1.\normalbaselineskip}[0pt][0pt]{\multirow{2}{*}{\rotatebox[origin=c]{90}{Stage 1}}} &
    \includegraphics[width=.11\textwidth]{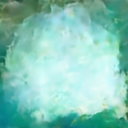} &
    \includegraphics[width=.11\textwidth]{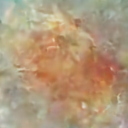} &
    \includegraphics[width=.11\textwidth]{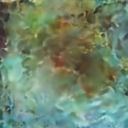} &
    \includegraphics[width=.11\textwidth]{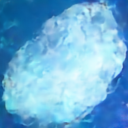} &
    \includegraphics[width=.11\textwidth]{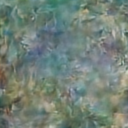} &
    \includegraphics[width=.11\textwidth]{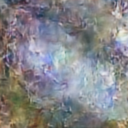} &
    \includegraphics[width=.11\textwidth]{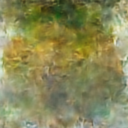} &
    \includegraphics[width=.11\textwidth]{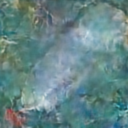} \\
    &
    \includegraphics[width=.11\textwidth]{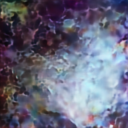} &
    \includegraphics[width=.11\textwidth]{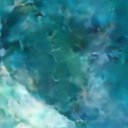} &
    \includegraphics[width=.11\textwidth]{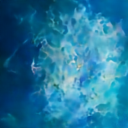} &
    \includegraphics[width=.11\textwidth]{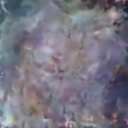} &
    \includegraphics[width=.11\textwidth]{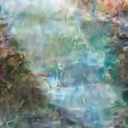} &
    \includegraphics[width=.11\textwidth]{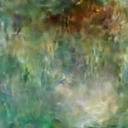} &
    \includegraphics[width=.11\textwidth]{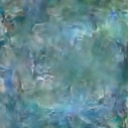} &
    \includegraphics[width=.11\textwidth]{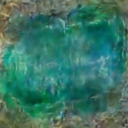}  \\
        \addlinespace[0.5cm]
      \raisebox{1.\normalbaselineskip}[0pt][0pt]{\multirow{2}{*}{\rotatebox[origin=c]{90}{Stage 2}}}&   
    \includegraphics[width=.11\textwidth]{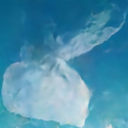} &
    \includegraphics[width=.11\textwidth]{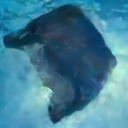} &
    \includegraphics[width=.11\textwidth]{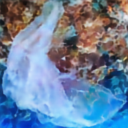} &
    \includegraphics[width=.11\textwidth]{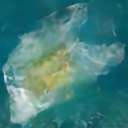} &
    \includegraphics[width=.11\textwidth]{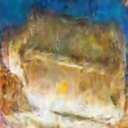} &
    \includegraphics[width=.11\textwidth]{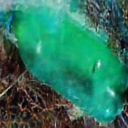} &
    \includegraphics[width=.11\textwidth]{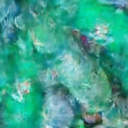} &
    \includegraphics[width=.11\textwidth]{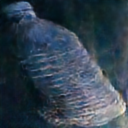}   \\
    
&
    \includegraphics[width=.11\textwidth]{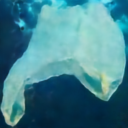} &
    \includegraphics[width=.11\textwidth]{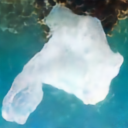} &
    \includegraphics[width=.11\textwidth]{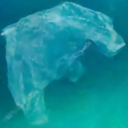} &
    \includegraphics[width=.11\textwidth]{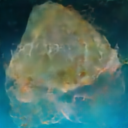}  &
    \includegraphics[width=.11\textwidth]{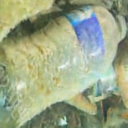} &
    \includegraphics[width=.11\textwidth]{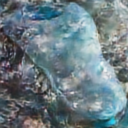} &
    \includegraphics[width=.11\textwidth]{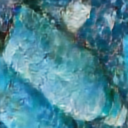} &
    \includegraphics[width=.11\textwidth]{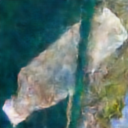}  \\
    \addlinespace[0.5cm]
      \raisebox{2\normalbaselineskip}[0pt][0pt]{\multirow{2}{*}{\rotatebox[origin=c]{90}{Reconstruction}}}&
    \includegraphics[width=.11\textwidth]{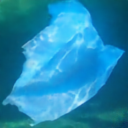} &
    \includegraphics[width=.11\textwidth]{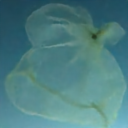} &
    \includegraphics[width=.11\textwidth]{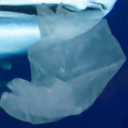} &
    \includegraphics[width=.11\textwidth]{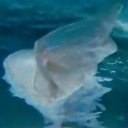} &
    \includegraphics[width=.11\textwidth]{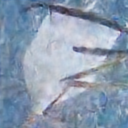} &
    \includegraphics[width=.11\textwidth]{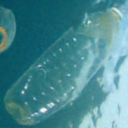} &
    \includegraphics[width=.11\textwidth]{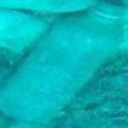} &
    \includegraphics[width=.11\textwidth]{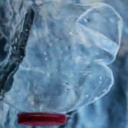}   \\
    
    &
    \includegraphics[width=.11\textwidth]{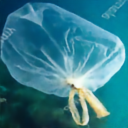} &
    \includegraphics[width=.11\textwidth]{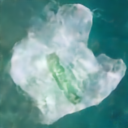} &
    \includegraphics[width=.11\textwidth]{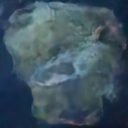} &
    \includegraphics[width=.11\textwidth]{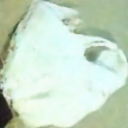} &
    \includegraphics[width=.11\textwidth]{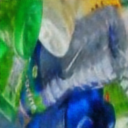} &
    \includegraphics[width=.11\textwidth]{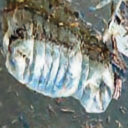} &
    \includegraphics[width=.11\textwidth]{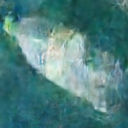} &
    \includegraphics[width=.11\textwidth]{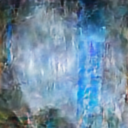} \\
  \end{tabular}
  \caption{Sample generated images from the two-stage VAE for each trash class. }
  \label{fig:plasticbag}
\end{figure*}

\subsection{Binary Classifier}
A binary classifier was built on ResNet-50 \cite{he2016deep} to select good quality images from the generated images. The original input dimension of ResNet-50 was $(224,224,3)$, but we changed to $(128,128,3)$ to match the input dimension to the output dimension of the two-stage VAE model. The remaining structure of ResNet-50 was maintained. 

\subsection{Multi-class Classifier}
The network for a multi-class classifier was designed to quantify the improvements to object classification tasks using generated images. The input is resized to $(32,32,3)$. It has two convolutional layers followed by one dense layer and one dropout layer. A softmax function is used as the activation function for the output layer and classifies three classes. One class is for our generated images, and the other two are underwater background and fish. The architecture of the network is shown in Fig. \ref{fig:nn}.

We use three metrics to evaluate the performance of the multi-class classifier: precision, recall and F1 score. Precision is useful if the cost of false positive is high. Recall is important when the cost of false negative is high. Lastly, F1 score provides the balanced information between precision and recall \cite{goutte2005probabilistic}. In our case of classifying trash, the cost of false positives is relatively higher than the cost of false negatives. This is because the consequences of accidentally removing something from the environment that is not trash (i.e. fish, coral, etc.) are much worse than the consequences of missing a single piece of trash.

%For instance, when an AUV is collecting underwater trash, it is more important to have a high confidence level of trash classification.

\section{Experiments and Results}
\label{sec:experiments}
\begin{figure*}[h]
\setlength{\lineskip}{0pt}
\centering
\setlength\tabcolsep{1.pt}
\renewcommand{\arraystretch}{0.5}
  \begin{tabular}{c@{\extracolsep{0.1cm}}c@{\extracolsep{0.1cm}}c@{\extracolsep{0.1cm}}c@{\extracolsep{0.1cm}}c @{\extracolsep{0.5cm}} c@{\extracolsep{0.1cm}}c@{\extracolsep{0.1cm}}c@{\extracolsep{0.1cm}}c}
  & \multicolumn{4}{c}{Good quality} & \multicolumn{4}{c}{Bad quality} \\
  \raisebox{2.\normalbaselineskip}[0pt][0pt]{\rotatebox[origin=c]{90}{Bag}} &
    \includegraphics[width=.11\textwidth]{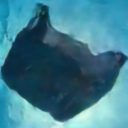} &
    \includegraphics[width=.11\textwidth]{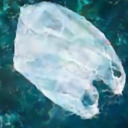} &
    \includegraphics[width=.11\textwidth]{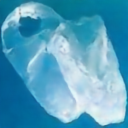} &
    \includegraphics[width=.11\textwidth]{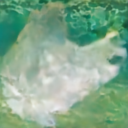} &
    \includegraphics[width=.11\textwidth]{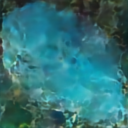} &
    \includegraphics[width=.11\textwidth]{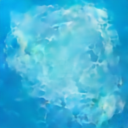} &
    \includegraphics[width=.11\textwidth]{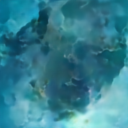} &
    \includegraphics[width=.11\textwidth]{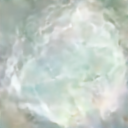}    \\
    \addlinespace[0.5cm]
  \raisebox{2.\normalbaselineskip}[0pt][0pt]{\rotatebox[origin=c]{90}{Bottle}} &
    \includegraphics[width=.11\textwidth]{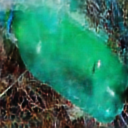} &
    \includegraphics[width=.11\textwidth]{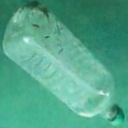} &
    \includegraphics[width=.11\textwidth]{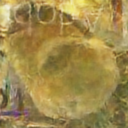} &
    \includegraphics[width=.11\textwidth]{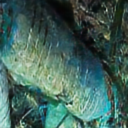} &
    \includegraphics[width=.11\textwidth]{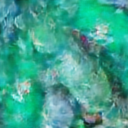} &
    \includegraphics[width=.11\textwidth]{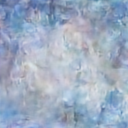} &
    \includegraphics[width=.11\textwidth]{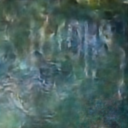} &
    \includegraphics[width=.11\textwidth]{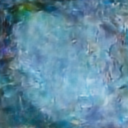} \\
  \end{tabular}
  \caption{Sample outputs from the binary classifier. ``Good quality'' generated images of each class are used to expand the dataset whereas the ``bad quality'' images are rejected. Qualitatively, it is apparent that ``bad quality'' images are poor representations of objects of their class. }
  \label{fig:binaryclass}
\end{figure*}

We trained our models on an NVIDIA Titan-class GPU with the Tensorflow library. One epoch takes 1-2 seconds for the stage 2 VAE, the binary classifier, and multi-class classifier. It takes 17-18 seconds
for each epoch of the stage 1 VAE model. 
\subsection{Data Collection}
From web scraping and the J-EDI dataset, we collected $775$ images of plastic bags and $283$ images of plastic bottles. We augmented each class of images by flipping horizontally, vertically, and rotating them by 90 degrees. As a result, we had a total of $3,000$ images of plastic bags and $1,000$ images of plastic bottles. For training and testing the multi-class classifier, we used the QUT fish dataset \cite{anantharajah2014local} which has $4,405$ fish images and randomly selected $3,000$ images. We also collected $271$ images of underwater scenes without objects and augmented them using the method described above, resulting in $3,000$ images.

\subsection{Image Generation}
Two two-stage VAE models are trained separately, one for each class of trash. We used different parameters to train each model due to the significant difference in the size of the collected datasets. During the training process, we use the mean absolute error (MAE) \cite{chai2014root} in Eq. \ref{eq:mae} as a stopping criterion for each stage of the VAE in addition to the loss functions from the original VAE. The MAE is used since it is robust to outliers and helps to evaluate the training progress. 

\begin{equation}\label{eq:mae}
    MAE = \frac{1}{n}\sum_{i=1}^{n}|x_{i} - \hat{x}_{i}| 
\end{equation}

\begin{enumerate}
    \item Plastic Bag: For the stage 1 VAE, we use ResNet with $16$ as its base dimension and $3$ for the kernel size, training it for $3,000$ epochs. For the stage 2 VAE, four layers of $1024$-dimension dense layers are used. The dimension of the latent variable is $12$ and the batch size is $16$. We train for $6,000$ epochs for the stage 2 VAE. 
    \item Plastic Bottle: For the stage 1 VAE, we use the same network and parameters as the model for the plastic bag class, also training for $3,000$ epochs. For the stage 2 VAE, four layers of $1024$-dimension dense layers are used. The dimension of the latent variable is $8$ and the batch size is $8$. We train for $6,000$ epochs for the stage 2 VAE.
\end{enumerate}
% These hyperparameters are determined empirically through a series of experiments.

The samples of generated images are shown in Fig. \ref{fig:plasticbag}, and Table \ref{tb:fid} presents the FID scores for images generated from stage 1, stage 2, and reconstructed images. It is clear that the two-stage VAE improves the visual quality of generated images for both plastic bag and bottle classes. The images from stage 1 are barely distinctive between classes. On the other hand, the images from stage 2 are significantly sharper in general. %It is also notable that  quality of images for each class. 
Although the two-stage VAE for the plastic bottle class outputs decent results, the generated plastic bag images are more crisp. This is also observed from the FID score.
% \begin{table*}[t!]
%     \vspace{3mm}
%     \caption{Generation and classification evaluation}
%     \vspace{2mm}
%     \begin{subtable}{.5\linewidth}
%       \centering
%   \caption{FID scores for the two-stage VAE}
%   \begin{tabular}{p{1cm}|p{1.5cm}|p{1.5cm}|p{1.5cm}}
%      & Stage 1 & Stage 2 & Reconstruction \\ \hline
%      Bag & 98 & 196 & 175 \\ \hline
%      Bottle & 223 & 301 & 240 \\
%   \end{tabular}
%     \label{tb:fid}
%     \end{subtable}%
%     \begin{subtable}{.5\linewidth}
%       \centering
%  \caption{Accuracy for the binary classifier}
%   \begin{tabular}{p{1cm}|p{1.8cm}|p{1.8cm}|p{1.8cm}}
%      & Training Acc. & Validation Acc. & Test Acc. \\ \hline
%      Bag & 0.89 & 0.88 & 0.86 \\ \hline
%      Bottle & 0.88 & 0.83 & 0.82 \\
%   \end{tabular}
%     \label{tb:binary}
%     \end{subtable} 
% \end{table*}

\begin{table*}[t!]
      \centering
  \caption{FID scores for the two-stage VAE}
  \begin{tabular}{p{1cm}|p{1.5cm}|p{1.5cm}|p{2cm}}
     & Stage 1 & Stage 2 & Reconstruction \\ \hline
     Bag & 98 & 196 & 175 \\ \hline
     Bottle & 223 & 301 & 240 \\
  \end{tabular}
    \label{tb:fid}
\end{table*}

\begin{table*}[t!]
      \centering
 \caption{Accuracy for the binary classifier}
  \begin{tabular}{p{1cm}|p{2cm}|p{2.3cm}|p{1.8cm}}
     & Training Acc. & Validation Acc. & Test Acc. \\ \hline
     Bag & 0.89 & 0.88 & 0.86 \\ \hline
     Bottle & 0.88 & 0.83 & 0.82 \\
  \end{tabular}
    \label{tb:binary}
\end{table*}

\subsection{Binary Classification}
% \input{results/binary.tex}
% % \input{results/binaryimages.tex}
% \input{results/resultstable.tex}
The binary classifier employs the ResNet-50 architecture. We first generated $10,000$ images each of plastic bags and bottles. Then, we labeled images as either ``good" or ``bad" using the visual quality of each image. We then built two separate training datasets using $1,200$ good quality images and $1,200$ bad quality images, one dataset for plastic bags and one for plastic bottles. The batch size was set to 16 and the epochs for training the classifier was 50. The training results are shown in Table \ref{tb:binary}. Similar to the outputs of the two-stage VAE, the binary classifier for the plastic bag class performs better overall. Lastly, the images classified as ``good quality" are added to the existing dataset for each class. The sample outputs from the binary classifier images are shown in Fig. \ref{fig:binaryclass}.

\begin{table*}[t!]
\centering
\footnotesize
\caption{Evaluation of object classification tasks}
\setlength\belowcaptionskip{-0pt}
  \begin{tabular}{p{.6cm}|p{.9cm}|p{.9cm}|p{1.cm}|p{.9cm}|p{.9cm}|p{.9cm}|p{1.cm}|p{.9cm}|p{.9cm}|p{.9cm}|p{1.cm}|p{.9cm}}
  \hline
  \multicolumn{13}{c}{Plastic Bag} \\ \hline
  \multirow{2}{*}{}&\multicolumn{4}{c|}{Real} &\multicolumn{4}{c|}{Generated}& \multicolumn{4}{c}{Mixed} \\ \cline{2-13}
 & Precision & Recall & F1 score & Support & Precision & Recall & F1 score & Support & Precision & Recall & F1 score & Support \\ \hline
  bag   & 0.96& 0.62& 0.76& 300& \textbf{0.96}& \textbf{0.80} & \textbf{0.87} & 300 & 0.95 & 0.78 & 0.86 & 300 \\ \hline
  fish& 0.95 & 0.95 & 0.95 & 300 & 0.94 & 0.97 & 0.95 & 300 & 0.95 & 0.96  & 0.95 & 300 \\ \hline
  empty & 0.74& 0.99& 0.84 & 300 &0.87& 0.99& 0.93& 300& 0.85& 0.99& 0.92 &300 \\ \hline \hline
  avg/tot  & 0.88 & 0.86& 0.85& 900 & 0.92 & 0.92 & 0.92 & 900 & 0.92 & 0.91 & 0.91 & 900 \\ \hline \hline
  \multicolumn{13}{c}{Plastic Bottle} \\ \hline
     \multirow{2}{*}{}&\multicolumn{4}{c|}{Real} &\multicolumn{4}{c|}{Generated}& \multicolumn{4}{c}{Mixed} \\ \cline{2-13}
& Precision & Recall & F1 score & Support & Precision & Recall & F1 score & Support & Precision & Recall & F1 score & Support \\ \hline
  bottle & 0.93 & 0.78 & 0.85 & 300 & 0.93 & 0.78 & 0.85 & 300 & \textbf{0.97} & \textbf{0.80} & \textbf{0.87} & 300 \\ \hline
  fish& 0.89 & 0.94 & 0.91 & 300 & 0.83 & 0.96 &0.89 &300 &0.91 &0.96 &0.93 &300 \\ \hline
  empty & 0.89& 0.98&0.94 & 300&0.96 & 0.97 & 0.96& 300& 0.89 & 1.00 &0.94 & 300 \\ \hline \hline
  avg/tot & 0.90& 0.90&0.90 &900 &0.91 &0.90 &0.90 &900 &0.92 &0.92 &0.92 &900 \\ \hline \hline
\end{tabular}
\vspace{0mm}
\label{tb:resulteval}
\end{table*}
\subsection{Multi-class Classification}
To quantify the impact of the generated images, three different types of datasets are created as follows. Each dataset contains $3,000$ images so only the compositions of the datasets are different among them. 
\begin{enumerate}
    \item Real dataset: only includes the real images from the original data collection.
    \item Generated dataset: consists of the images which are generated from the two-stage VAE and filtered by the binary classifier.
    \item Mixed dataset: consists of 50\% images from the real dataset and 50\% from the generated dataset. 
\end{enumerate}

For each class, 3 multi-class classifiers are trained separately with each dataset. The batch size is 100, and the epoch is 30 for all training processes. Test images are 300 real images for each dataset, and the images are not shown to the classifiers during the training processes. In total, 6 multi-classifiers are trained, and the results are shown in Table \ref{tb:resulteval}. 

In both plastic bag and bottle classes, the classifiers trained with the generated dataset and mixed dataset outperform the ones trained with the real dataset in general. Recall and F1 score are improved up to $18\%$ and $11 \%$, respectively, in the plastic bag case. Precision becomes $4 \%$ more accurate in the plastic bottle case. For the underwater trash detection problem, therefore, the results point to the viability of using the proposed approach to augment an existing yet small dataset with generated imagery using a two-stage VAE.
%approach.

%\subsection{Discussion}
%Our proposed approach to expand a dataset is evaluated. 

\section{Conclusion}
We present an approach to address data scarcity problems in underwater image datasets for visual detection of marine debris. The proposed approach relies on a two-stage variational autoencoder (VAE) and a binary classifier to evaluate the generated imagery for quality and realism. From the images generated by the two-stage VAE, the binary classifier selects ``good quality'' images and augments the given dataset with them. Lastly, a multi-class classifier is used to evaluate the impact of the augmentation process by measuring the accuracy of an object detector trained on combinations of real and generated trash images. Our results show that the classifier trained with the augmented data outperforms the one trained only with the real data. This approach will not only be valid for the underwater trash classification problem presented in this paper, but it will also be useful for any data-dependent task for which collecting more images is infeasible or challenging at best. Future work will involve augmenting the existing underwater trash dataset further with real and generated data and releasing it to the broader research community, and embedding a multi-class trash detector trained with this expanded dataset on-board an actual AUV platform for field trials in open-water environments. 
% \input{results/plasticbag_v2.tex}
% \section*{Acknowledgments}
% We are thankful to Owen Queeglay, Kevin Orpen, and Kimberly Barthelemy for their assistance.

\bibliographystyle{abbrv}
\bibliography{citation}

\begin{thebibliography}{10}

\bibitem{anantharajah2014local}
K.~{Anantharajah}, {ZongYuan Ge}, C.~{McCool}, S.~{Denman}, C.~{Fookes},
  P.~{Corke}, D.~{Tjondronegoro}, and S.~{Sridharan}.
\newblock Local inter-session variability modelling for object classification.
\newblock In {\em IEEE Winter Conference on Applications of Computer Vision},
  pages 309--316, March 2014.

\bibitem{antoniou2017data}
A.~Antoniou, A.~Storkey, and H.~Edwards.
\newblock {Data augmentation generative adversarial networks}.
\newblock {\em arXiv preprint arXiv:1711.04340}, 2017.

\bibitem{arjovsky2017wasserstein}
M.~Arjovsky, S.~Chintala, and L.~Bottou.
\newblock {Wasserstein GAN}.
\newblock {\em arXiv preprint arXiv:1701.07875}, 2017.

\bibitem{brock2018large}
A.~Brock, J.~Donahue, and K.~Simonyan.
\newblock {Large scale GAN training for high fidelity natural image synthesis}.
\newblock {\em arXiv preprint arXiv:1809.11096}, 2018.

\bibitem{brock2016neural}
A.~Brock, T.~Lim, J.~M. Ritchie, and N.~Weston.
\newblock {Neural photo editing with introspective adversarial networks}.
\newblock {\em arXiv preprint arXiv:1609.07093}, 2016.

\bibitem{cao2018recent}
Y.-J. Cao, L.-L. Jia, Y.-X. Chen, N.~Lin, C.~Yang, B.~Zhang, Z.~Liu, X.-X. Li,
  and H.-H. Dai.
\newblock {Recent Advances of Generative Adversarial Networks in Computer
  Vision}.
\newblock {\em IEEE Access}, 7:14985--15006, 2018.

\bibitem{chai2014root}
T.~Chai and R.~R. Draxler.
\newblock {Root mean square error (RMSE) or mean absolute error
  (MAE)?--Arguments against avoiding RMSE in the literature}.
\newblock {\em Geoscientific model development}, 7(3):1247--1250, 2014.

\bibitem{chen2017multi}
X.~Chen, H.~Ma, J.~Wan, B.~Li, and T.~Xia.
\newblock {Multi-view 3D object detection network for autonomous driving}.
\newblock In {\em Proceedings of the IEEE Conference on Computer Vision and
  Pattern Recognition}, pages 1907--1915, 2017.

\bibitem{dai2019diagnosing}
B.~Dai and D.~Wipf.
\newblock {Diagnosing and enhancing VAE models}.
\newblock {\em arXiv preprint arXiv:1903.05789}, 2019.

\bibitem{fabbri2018enhancing}
C.~Fabbri, M.~J. Islam, and J.~Sattar.
\newblock {Enhancing underwater imagery using generative adversarial networks}.
\newblock In {\em 2018 IEEE International Conference on Robotics and Automation
  (ICRA)}, pages 7159--7165. IEEE, 2018.

\bibitem{friedman1997bayesian}
N.~Friedman, D.~Geiger, and M.~Goldszmidt.
\newblock {Bayesian network classifiers}.
\newblock {\em Machine learning}, 29(2-3):131--163, 1997.

\bibitem{fulton2018robotic}
M.~{Fulton}, J.~{Hong}, M.~J. {Islam}, and J.~{Sattar}.
\newblock {Robotic Detection of Marine Litter Using Deep Visual Detection
  Models}.
\newblock In {\em 2019 International Conference on Robotics and Automation
  (ICRA)}, pages 5752--5758, May 2019.

\bibitem{garcia2017few}
V.~Garcia and J.~Bruna.
\newblock {Few-shot learning with graph neural networks}.
\newblock {\em arXiv preprint arXiv:1711.04043}, 2017.

\bibitem{goodfellow2014generative}
I.~Goodfellow, J.~Pouget-Abadie, M.~Mirza, B.~Xu, D.~Warde-Farley, S.~Ozair,
  A.~Courville, and Y.~Bengio.
\newblock {Generative adversarial nets}.
\newblock In {\em Advances in Neural Information Processing Systems}, pages
  2672--2680, 2014.

\bibitem{goutte2005probabilistic}
C.~Goutte and E.~Gaussier.
\newblock {A probabilistic interpretation of precision, recall and F-score,
  with implication for evaluation}.
\newblock In {\em European Conference on Information Retrieval}, pages
  345--359. Springer, 2005.

\bibitem{guvenc2018detection}
I.~Guvenc, F.~Koohifar, S.~Singh, M.~L. Sichitiu, and D.~Matolak.
\newblock {Detection, tracking, and interdiction for amateur drones}.
\newblock {\em IEEE Communications Magazine}, 56(4):75--81, 2018.

\bibitem{he2016deep}
K.~He, X.~Zhang, S.~Ren, and J.~Sun.
\newblock {Deep residual learning for image recognition}.
\newblock In {\em Proceedings of the IEEE Conference on Computer Vision and
  Pattern Recognition}, pages 770--778, 2016.

\bibitem{heusel2017gans}
M.~Heusel, H.~Ramsauer, T.~Unterthiner, B.~Nessler, and S.~Hochreiter.
\newblock {GANs trained by a two time-scale update rule converge to a local
  Nash equilibrium}.
\newblock In {\em Advances in Neural Information Processing Systems}, pages
  6626--6637, 2017.

\bibitem{JAMSTECDebri}
{\relax Japan Agency for Marine Earth Science and Technology}.
\newblock {\em {Deep-sea Debris Database}}, 2018.
\newblock {\small
  \url{http://www.godac.jamstec.go.jp/catalog/dsdebris/e/index.html}}. Accessed
  02-10-2018.

\bibitem{kingma2016improved}
D.~P. Kingma, T.~Salimans, R.~Jozefowicz, X.~Chen, I.~Sutskever, and
  M.~Welling.
\newblock {Improved variational inference with inverse autoregressive flow}.
\newblock In {\em Advances in Neural Information Processing Systems}, pages
  4743--4751, 2016.

\bibitem{kingma2013auto}
D.~P. Kingma and M.~Welling.
\newblock {Auto-encoding variational Bayes}.
\newblock {\em arXiv preprint arXiv:1312.6114}, 2013.

\bibitem{li2017watergan}
J.~Li, K.~A. Skinner, R.~M. Eustice, and M.~Johnson-Roberson.
\newblock {WaterGAN: Unsupervised generative network to enable real-time color
  correction of monocular underwater images}.
\newblock {\em IEEE Robotics and Automation letters}, 3(1):387--394, 2017.

\bibitem{mirza2014conditional}
M.~Mirza and S.~Osindero.
\newblock {Conditional generative adversarial nets}.
\newblock {\em arXiv preprint arXiv:1411.1784}, 2014.

\bibitem{palatucci2009zero}
M.~Palatucci, D.~Pomerleau, G.~E. Hinton, and T.~M. Mitchell.
\newblock {Zero-shot learning with semantic output codes}.
\newblock In {\em Advances in Neural Information Processing Systems}, pages
  1410--1418, 2009.

\bibitem{perez2017effectiveness}
L.~Perez and J.~Wang.
\newblock {The effectiveness of data augmentation in image classification using
  deep learning}.
\newblock {\em arXiv preprint arXiv:1712.04621}, 2017.

\bibitem{salimans2016improved}
T.~Salimans, I.~Goodfellow, W.~Zaremba, V.~Cheung, A.~Radford, and X.~Chen.
\newblock {Improved techniques for training GANs}.
\newblock In {\em Advances in Neural Information Processing Systems}, pages
  2234--2242, 2016.

\bibitem{szegedy2015going}
C.~Szegedy, W.~Liu, Y.~Jia, P.~Sermanet, S.~Reed, D.~Anguelov, D.~Erhan,
  V.~Vanhoucke, and A.~Rabinovich.
\newblock {Going deeper with convolutions}.
\newblock In {\em Proceedings of the IEEE Conference on Computer Vision and
  Pattern Recognition}, pages 1--9, 2015.

\bibitem{tolstikhin2017wasserstein}
I.~Tolstikhin, O.~Bousquet, S.~Gelly, and B.~Schoelkopf.
\newblock {Wasserstein auto-encoders}.
\newblock {\em arXiv preprint arXiv:1711.01558}, 2017.

\bibitem{torrey2010transfer}
L.~Torrey and J.~Shavlik.
\newblock {Transfer learning}.
\newblock In {\em Handbook of research on machine learning applications and
  trends: algorithms, methods, and techniques}, pages 242--264. IGI Global,
  2010.

\bibitem{wang2019generative}
Z.~Wang, Q.~She, and T.~E. Ward.
\newblock {Generative Adversarial Networks: A Survey and Taxonomy}.
\newblock {\em arXiv preprint arXiv:1906.01529}, 2019.

\bibitem{wisse2018milking}
D.-J. Wisse, P.~J.~M. Van~Adrichem, and K.~Van~den Berg.
\newblock {Milking robot with kick detection}, Dec.~6 2018.
\newblock US Patent App. 15/761,593.

\bibitem{zhang2018deep}
L.~Zhang, J.~Jia, G.~Gui, X.~Hao, W.~Gao, and M.~Wang.
\newblock {Deep learning based improved classification system for designing
  tomato harvesting robot}.
\newblock {\em IEEE Access}, 6:67940--67950, 2018.

\bibitem{zhu2017unpaired}
J.-Y. Zhu, T.~Park, P.~Isola, and A.~A. Efros.
\newblock {Unpaired image-to-image translation using cycle-consistent
  adversarial networks}.
\newblock In {\em Proceedings of the IEEE International Conference on Computer
  Vision}, pages 2223--2232, 2017.

\bibitem{zivkovic2004improved}
Z.~Zivkovic.
\newblock {Improved adaptive Gaussian mixture model for background
  subtraction}.
\newblock In {\em Proceedings of the 17th International Conference on Pattern
  Recognition, 2004. ICPR 2004.}, volume~2, pages 28--31. IEEE, 2004.

\end{thebibliography}
\end{document}